\documentclass[journal]{IEEEtran}
\usepackage{epsf,epsfig,subfig, amsmath, amssymb, bm,amsbsy,float,booktabs}
\usepackage[utf8]{inputenc}
\usepackage{xcolor}
\usepackage{hyperref}
\hypersetup{
    colorlinks=true,
    linkcolor=red,
    urlcolor=blue,
    citecolor=red,
    citebordercolor=red,
    linkbordercolor=red,
    filebordercolor=red
}
\usepackage[english]{babel}
\usepackage{tikz}
\usepackage{cite}
\usepackage{lineno}
\usetikzlibrary{shapes,arrows}
\usepackage{algorithm}
\usepackage{algpseudocode}
\graphicspath{ {./Images/} }
\renewcommand\IEEEkeywordsname{Keywords}
\algnewcommand{\Initialize}[1]{%
  \State \textbf{Initialize [Radio:0]:}
  \Statex \hspace*{\algorithmicindent}\parbox[t]{.8\linewidth}{\raggedright #1}
}

\begin{document}
\title{Three-Way Deep Neural Network for Radio Frequency Map Generation and Source Localization}
\author{Kuldeep S. Gill, Son Nguyen, Myo M. Thein, Alexander M. Wyglinski\\
\normalsize $^\dag$Department of Electrical and Computer Engineering, Worcester Polytechnic Institute, Worcester, MA\\
\normalsize \{~ksgill,~snguyen,~mmthein,~alexw\}@wpi.edu}

\maketitle
\begin{abstract}
In this paper, we present a Generative Adversarial Network (GAN) machine learning model to interpolate irregularly distributed measurements across the spatial domain to construct a smooth radio frequency map (RFMap) and then perform localization using a deep neural network. Monitoring wireless spectrum over spatial, temporal, and frequency domains will become a critical feature in facilitating dynamic spectrum access (DSA) in beyond-5G and 6G communication technologies. Localization, wireless signal detection, and spectrum policy-making are several of the applications where distributed spectrum sensing will play a significant role. Detection and positioning of wireless emitters is a very challenging task in a large spectral and spatial area.  In order to construct a smooth RFMap database, a large number of measurements are required which can be very expensive and time consuming. One approach to help realize these systems is to collect finite localized measurements across a given area and then interpolate the measurement values to construct the database. Current methods in the literature employ channel modeling to construct the radio frequency map, which lacks the granularity for accurate localization whereas our proposed approach reconstructs a new generalized RFMap. Localization results are presented and compared with conventional channel models.
\end{abstract}
\vspace{2pt}
\begin{IEEEkeywords}
\textbf{Generative Adversarial Network, Dynamic Spectrum Access, localization, 6G}
\end{IEEEkeywords}

\section{Introduction}
Cellular networks technologies are rapidly evolving towards self-organizing networks (SON) via the integration of artificial intelligence and emerging wireless technologies~\cite{dottling2009challenges}. Accurate measurement and estimation of the radio frequency (RF) environment has become increasingly important for supporting SON operations. Although theoretical models such as COST231~\cite{damosso1999cost} or Okumura-Hata~\cite{hata1980empirical} are extensively used by the wireless community to understand the RF propagation environment, they do not fully characterize the effects of absorption, reflection, and refraction, thus yielding inaccurate RF estimates especially in urban environments with dense population distributions~\cite{low1992comparison}. In particular,  it has been demonstrated that opportunistic RF estimation using mobile devices has yielded promising results~\cite{schuette2015performance, wu20203d}. Additionally, based on measurements obtained from mobile devices employed across a geographical region. The estimation and prediction of RF characteristics can be performed for areas with no mobile devices present by employing interpolation techniques.   

\begin{figure}[!ht]
    \centering
    \includegraphics[width=0.475\textwidth]{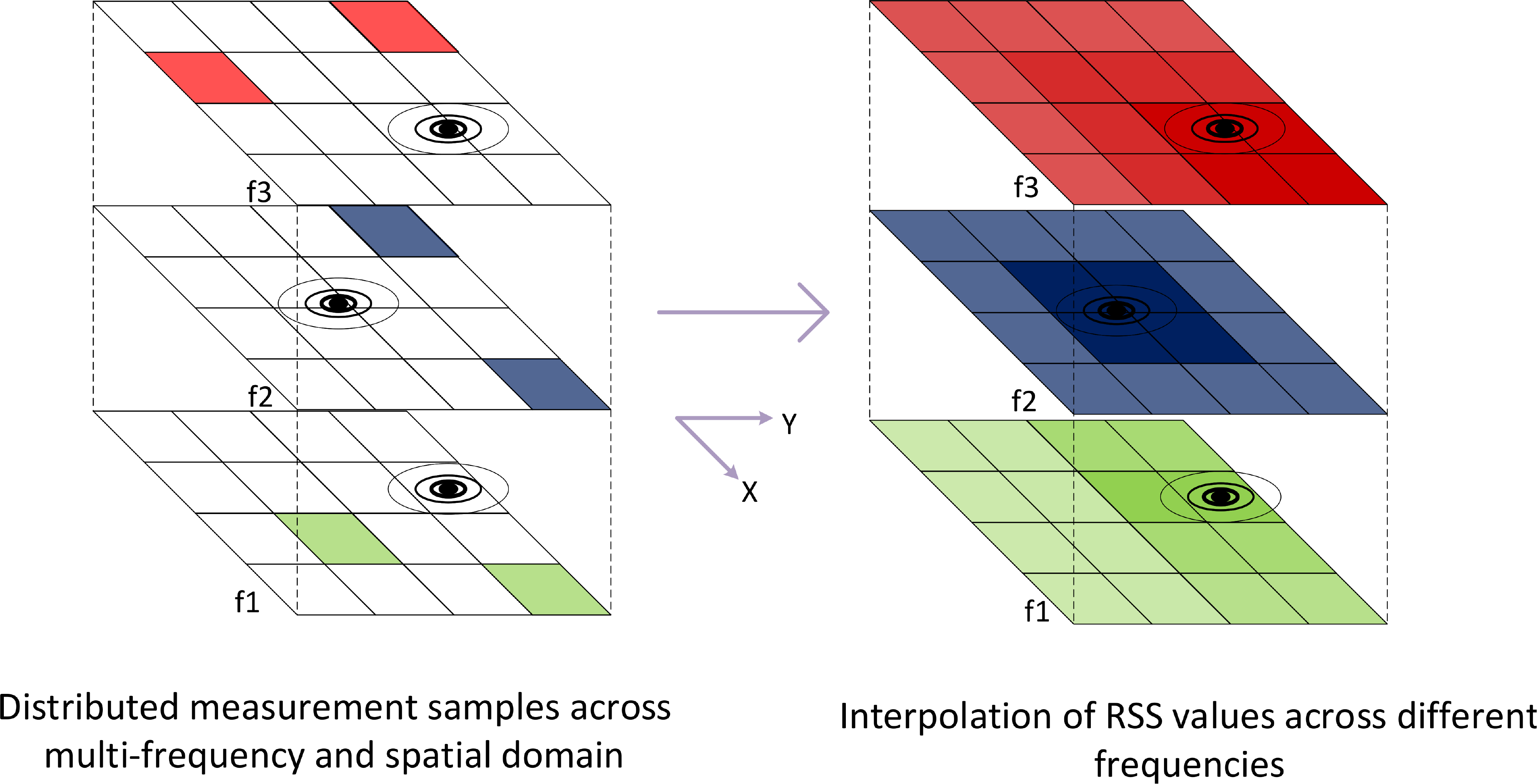}
    \caption{RSS values for three emitters at different frequencies which are interpolated to create a smooth RFMap database using generative model.}
    \label{fig:rfmap}
    \vspace{-0.65cm}
\end{figure}

RF-based localization using fingerprint mapping requires an initial training step, where a database consisting of receive signal strength indicator (RSSI) values is constructed. This database is constructed by measuring the RSSI values at specific locations across a localization area using a transceiver. The accuracy of the localization depends on the measured values and their quantity. Traditionally, both Okumura-Hata and COST-231 have been used to estimate the RF environment, which are based on a collection of theoretical computations~\cite{shabbir2011comparison, gajewski2018propagation}. In Refernce~\cite{ezpeleta2015rf}, the authors have explored different interpolation functions for constructing a smoother RFMap for enhancing the localization accuracy. Nearest neighbor (NN) interpolation is one of the most widely used data-driven methods, where the unmeasured location values are filled with the mean of their nearby observations to construct the RFMap for localization~\cite{xie2016improved, guowei2013research}. 

However, employing $k$-NN and propagation channel models for localization have several issues when employed in urban scenarios, such as:
\begin{itemize}
    \item Constructing a database possessing a large number of values can yield more accurate localization estimates but it is expensive and time-consuming.
    \item Field tests have shown that propagation models are potentially prone to failure with respect to fine-grained localization accuracy.
    \item The $k$-NN performs better relative to traditional propagation channel models but the accuracy degrades in dense urban environments.
\end{itemize}

In this paper, we employ RFMaps generated via three-way deep neural network (DNN) techniques to perform localization. The first and second DNNs are used in synergy to generate new training distribution through two-player game-theoretic approach and it is known as generative adversarial learning. Authors in~\cite{li2019sparsely} have employed self-supervising generative adversarial network (SS-GAN) in order to construct the new generalized RFMap and demonstrate the robustness of this technique. The third DNN in our paper is employed in order to perform localization on the output RFMap creating using the generator network. Our key contributions in this work are summarized as follows:
\begin{itemize}
    \item For constructing the initial training data-set, we have collected a small amount of uncorrelated samples to keep the measurement costs minimal. Measurements were collected inside a controlled laboratory environment using wireless local area network (WLAN) RSS values.
    \item Instead of using propagation models, we have employed generative model in order to construct a smooth RFMap database. We applied the proposed GAN model to construct a RFMap of a geographical region using a initial RSS values from a database. The smooth RFMap is subsequently applied to a deep neural network for source localization.
    \item In our proposed approach, we sample from the measurement data space in order to create new training dataset via SS-GAN, which is later used for localization. We have computed the localization errors using $k$-NN and Multiple Imputation by Chained Equations (MICE) methods and compared it with our proposed method and we observed 90.27\% and 53.19\% reduction in error, respectively.
\end{itemize}

The rest of the paper is organized as follows: In Section~\ref{sec:sec2}, we talk about RFMap and the conventional methods used for interpolation. In Section~\ref{sec:sec3}, we present the GAN and deep neural network architecture employed for RFMap generation and source localization respectively. Our custom-built test-bed setup is presented in Section~\ref{sec:sec4} along with the results, which are compared with tagged localization values. Finally, we conclude the paper with Section~\ref{sec:concl}, which summarizes the work and provides future direction for the research.

\section{RFMap Concept}
\label{sec:sec2}

There are several approaches through which a RFMap can be constructed and its accuracy depends on the type of method used. Channel modeling has been a very popular method until recently, where the RF signal strength is estimated due to large-scale (propagation loss) and small-scale (multipath) channel effects. One prevalent method of modeling large-scale propagation relies on the Friis equation~\cite{friis1946} but the accuracy of RSS estimation degrades in a dense channel environment. Consequently, crowd-sourcing approaches have been gaining popularity due to the low cost associated with data collection. Data collection is performed by sharing radio frequency values, which may not be accurate due to noisy sample locations~\cite{jiang2016probabilistic}. Professional site surveys are the best solution as they use expensive equipment to guarantee a high level of accuracy. There is still an issue of time, where the areas need to be covered and data collection is required at large number of locations. One inexpensive alternative is to collect the data from several locations and then interpolate the data to cover the entire geographical region creating smooth RFMap. The accuracy of RSS values for a smooth RFMap  depends on the interpolation technique used. In~\cite{vanhoy2013spatial}, the authors have employed Discrete Cosine Transform (DCT) for the interpolation of RSS values to construct a smooth RFMap:
\begin{equation}
    \mathrm{ X_{k1,k2} = \sum_{n_1=0}^{N_1 - 1} \sum_{n_2=0}^{N_2 - 1} x_{n1,n2}cos[w_{n_1}k_1]cos[w_{n_2}k_2]}
    \label{eq:2}
\end{equation}
where $X$ is the significant coefficient, $w_{n1}=\dfrac{\pi}{N_1}(n+\dfrac{1}{2})$, $x$ is the distance from the transmitter and $k$ is the data point index. They looked at the estimation of the RFMap as a sampling problem and kept the number of points to be sampled to a minimum for practical purposes to reduce the sampling cost, since a real-world system will sample the RFMap using a network of expensive spectrum sensors.  The authors in~\cite{denkovski2012reliability} looked at implementing a flexible REM design for a robust estimation of interference and coverage characteristics of wireless system using small number of measurement samples. Their approach is based on Inverse Distance Weighting (IDW) spatial interpolation, and an interpolation error of less than $10$ dB was achieved. Following the IDW method, the interpolation value for an arbitrary spatial point ($x$, $y$) is calculated using:
\begin{equation}
    \mathrm{ f(x,y) = \dfrac{\sum_{k=1}^{N} W_k(x,y)Q_k(x,y)} {\sum_{k=1}^{N} W_k(x,y)}}
    \label{eq:3}
\end{equation}
where $Q_k$ is the output of the nodal function of the data point $k$ and $W_k$ is the weight assigned to the referred neighboring point at location $(x,y)$. In this paper, we are using a GAN to construct an accurate RFMap using a limited number of measurement samples. Measurement samples are interpolated using $k$-Nearest Neighbor ($k$-NN) and MICE. After analyzing the performance of both the methods, we selected the MICE interpolation for our GAN model as the ground truth. Figure~\ref{fig:rfmap} shows the interpolation of RSS values across multiple frequency values with limited number of samples. For this work, we are only using RSS values interpolation across single frequency particularly $2.4$ GHz as this is the operating frequency for Wi-Fi devices. Multi-frequency interpolation will be the focus of future studies on this topic.

\section{Proposed GAN Framework for RFMap Generation}
\label{sec:sec3}

GANs ~\cite{goodfellow2014generative} are a type of deep learning technique that is increasingly being used for data augmentation. GANs originally belong to the field of unsupervised learning as they do not require the labels or response variable for the generation of the synthetic data. They are generative models that learn the underlying distribution or structure of the data without specifying the target value. GANs learn the intrinsic distribution of the classes in the dataset that has multiple classes, such that they generate the synthetic data for all the classes that belong to the original dataset, so long as we have large number of samples belonging to each of the classes. The two most commonly used metrics for the comparison of GAN performance are the Kullback-Leibler (KL) divergence and the Jensen Shannon (JS) divergence~\cite{goodfellow2014generative}. The KL divergence measure how a probability distribution $P$ diverges from another probability distribution $Q$. It is given by:

\begin{equation}
    \mathrm{KL}(Q||P) = Q(f)\log \bigg(\dfrac{Q(f)}{P(f)}\bigg).
    \label{eq:4}
\end{equation}
 
The KL divergence is not symmetric, whereas the JS divergence is symmetric. The locations where the distribution of $P(x)$ is zero and $Q(x)$ is non-zero, the effect is disregarded. This could be problematic when we have two distributions to be compared that are equally important. It is important that for the regions where $P(x)$ has a non null mass, $Q(x)$ also has a non-null mass. There is an integral term in the equation for the KL divergence which explains that if the distribution $Q(x)$ is chosen to minimize the KL metric, it is unlikely that $Q(x)$ will assign a lot of mass to regions where $P(x)$ is close to zero. Conversely, JS Divergence behaves in a similar manner for small values of $P(x)$ or $Q(x)$.

\subsection{Network Architecture}

The framework presented in Figure~\ref{fig:rfmap} illustrates the architecture of a GAN model. There exists two networks, namely, the discriminator network and the generator network. The generator network takes in random noise as input and generates data that follows the distribution of the real data. On the other hand, the discriminator network plays the role of a critic in which it discriminates between the real and the generated samples. The feedback provided by the discriminator network to the generator network helps it generate samples that are close to the real data. The input to the discriminator network is real data as well as the synthetic data generated by the generator network. The task is to compare the samples provided to the discriminator network to the real data. The discriminator network then classifies the sample as real or generated by providing an output probability between $0$ and $1$. Here, a $0$ is defined as the sample being fake and a $1$ indicates sample is real. Anything in between gives us a probability the samples is real. Depending on the outcome of the discriminator, both the networks try to optimize their own parameters by fine tuning their networks and becoming better at their objective.

\begin{figure}[!ht]
    \centering
    \includegraphics[width=0.475\textwidth, height=8cm, keepaspectratio]{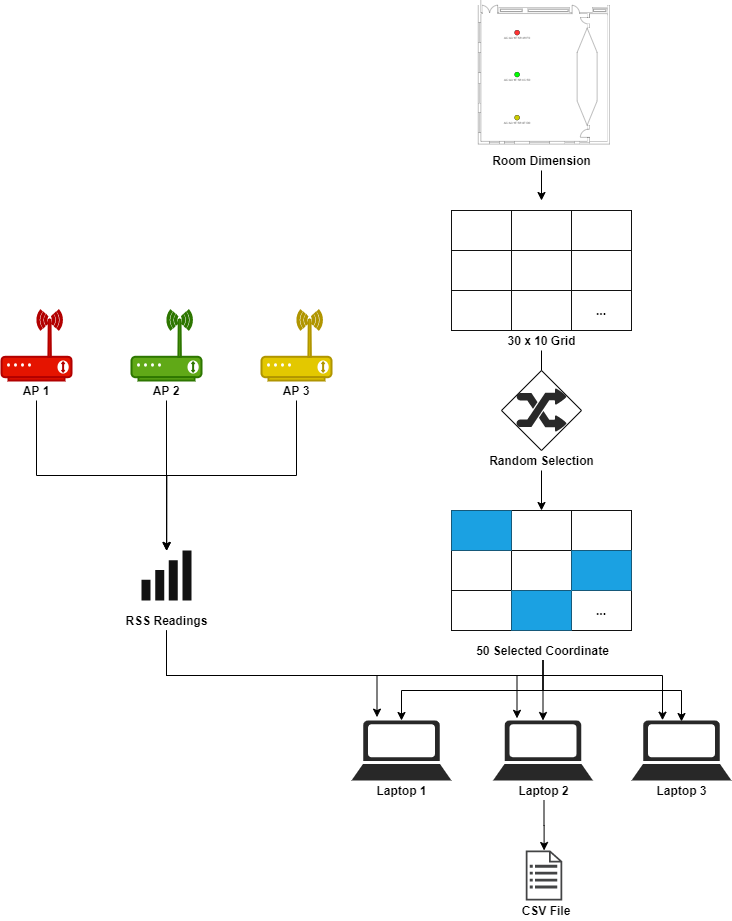}
    \caption{RSS data collected through three individual access points located in an indoor room by converting into rectangular grid.}
    \label{fig:rfmap}
    \vspace{-0.65cm}
\end{figure}

\begin{figure*}
    \centering
    \includegraphics[width=\textwidth]{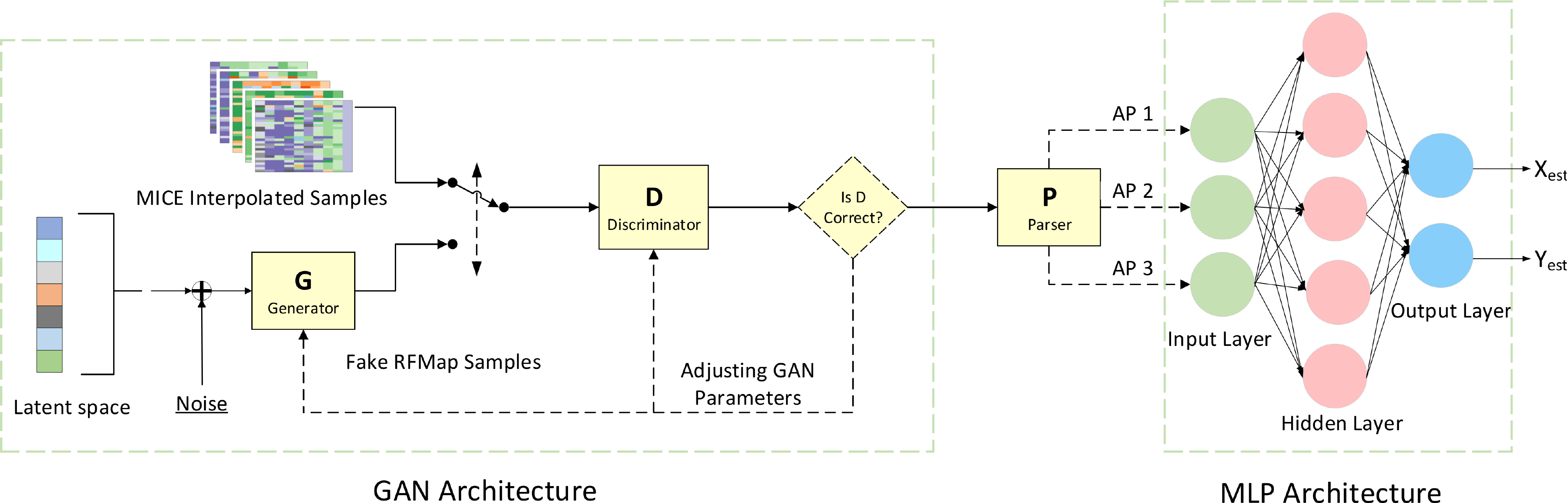}
    \caption{GAN Architecture to construct a smooth RFMap with a third deep neural network for emitter localization. The data-set generated by GAN is fed to third neural network for localization.}
    \label{fig:rfmap}
    \vspace{-0.75cm}
\end{figure*}

\subsection{Training GAN Model}

 If the discriminator is able to discriminate between the real and generated sample, the generator performs well in terms of creating artificial sample, making it more difficult for the discriminator to be able to distinguish between them in the next iteration. After training for a sufficiently long period of time, there would come a time when the discriminator outputs a probability of $0.5$, meaning that it is no longer able to distinguish between the real and synthetic data. This would be the ideal situation meaning the generator is producing realistic data such that the discriminator is not able to tell it apart from the real data. At this point, both networks cannot be improved anymore and have converged. This is a mini-max game with the value function $V(G, D)$, which is given by:
\begin{equation}
\begin{split}
     \min_{G}, \max_{D} V(D,G) &= E_{x~p_{data}(x)}[\log D(x)] \\
                               &+  E_{x~p_{z}(z)}[\log (1-D(G(z)))]
\end{split}
\label{eq:5}
\end{equation}
where $x$ is the real sample, $D(x)$ is the output of discriminator network, and $G(z)$ is output of generator. It has two loops, where the outer loop is trying to minimize the equation with respect to the generators parameters only and the inner loop is trying to maximize the equation with respect to the discriminator’s parameters only. Based on the log values, we observed the two networks are in an adversarial mode where they have opposing tasks in the game which they try to satisfy until convergence. Furthermore, if the discriminator is not able to classify the real and synthetic samples, it will update its parameters in the next iteration. The total reward for the discriminator is the total number of correct predictions that it makes while the reward for the generator is the total number of errors from the discriminator. This process continues until the parameters are optimized and equilibrium is achieved. Moreover, the discriminator weights are updated in such a manner that they maximize the probability of a real data sample $x$ being classified as belonging to the real data set. Conversely, the discriminator minimizes the probability that a fake sample is classified as belonging to the real data set. The loss or error function used maximizes the function $D(x)$ and also minimizes the $D(z)$ and $G(z)$. The log probability is used in the loss functions instead of raw probabilities since a log loss heavily penalizes the incorrect classifications of an algorithm that is confident about its predictions.

\section{Measurements and Experiment Results}
\label{sec:sec4}

This section describes the experimental setup used in the data collection process for this paper. Specifications regarding the enclosed area chosen for this experiment as well as the Wi-Fi access points used will be presented. Furthermore, the data collection routine and procedure we employed will also be described. Although the experimental setupemployed is specific to this paper without loss ingenerality, it can be readily replicated, modified, and adapted to collect data in different environment and scenarios. The first step is to measure the dimensions of the room and based on those measurements divide the room into grids. Alternatively, if the floor-plan of the designated experimental area is available, the length and width included in the floor-plan can be used instead of obtaining measurements. After the dimensions of the grid have been determined, a set number of points within the grid should be randomly selected. Depending on the size of the selected area, the number of randomly selected points ranges from $50$ to $200$ points. At each of the points, approximately $100$ to $200$ RSS readings are recorded for each of the access points. Finally, with data collected the RSS readings were stored in a \emph{csv} files for post-processing. 

\subsection{Experimental Setup}
For the purpose of this experiment, Room $116$ in the Atwater Kent Laboratories building at Worcester Polytechnic Institute (125 Salisbury St, Worcester, MA, 01609, USA) was selected since it was a large open indoor location with multiple access points distributed across the whole area of the room. The experiment area is $10.75$ m x $17.4$ m, which is divided into $30$ x $10$ grids. The area and dimension of the grid was decided based on the seating area in distributed lecture hall. The model of the access point used in the experiment was the Aruba $310$ \cite{aruba310}, which supports both IEEE $802.11$ac and $802.11$n. For the purpose of this experiment, traffic of these access points IEEE $802.11$n were collected. The access points used omni-directional antennas with a maximum transmit power of $+21$ dBm at $2.4$ GHz and $+24$ dBm at $5$ GHz. There are many access points that border the designated experimental area but there are only three access points that are physically inside the room. Regarding the data collection process, $50$ points within the grid are randomly selected using a custom Python script. At each point, the RSS readings are collected using the Homedale Wi-Fi monitor software for a duration of five minutes. Three laptops were used to collect the RSS readings with the $50$ points divided among the three laptops. In total, the data collection process yielded roughly $500$ data points, which were stored in \textit{.csv} files for input into the GAN training program.

Figure~\ref{fig:rfmap} describes the information flow in our proposed three-party DNN. The generator provides data to the discriminator, which then attempts to detect whether an observation is synthetic or real.  The inverse map (localization DNN model) is then trained on a RFMap images generated using GAN with $X$ and $Y$ coordinate values as the regression output. 

\begin{figure}[!ht]
    \centering
    \includegraphics[width=0.475\textwidth]{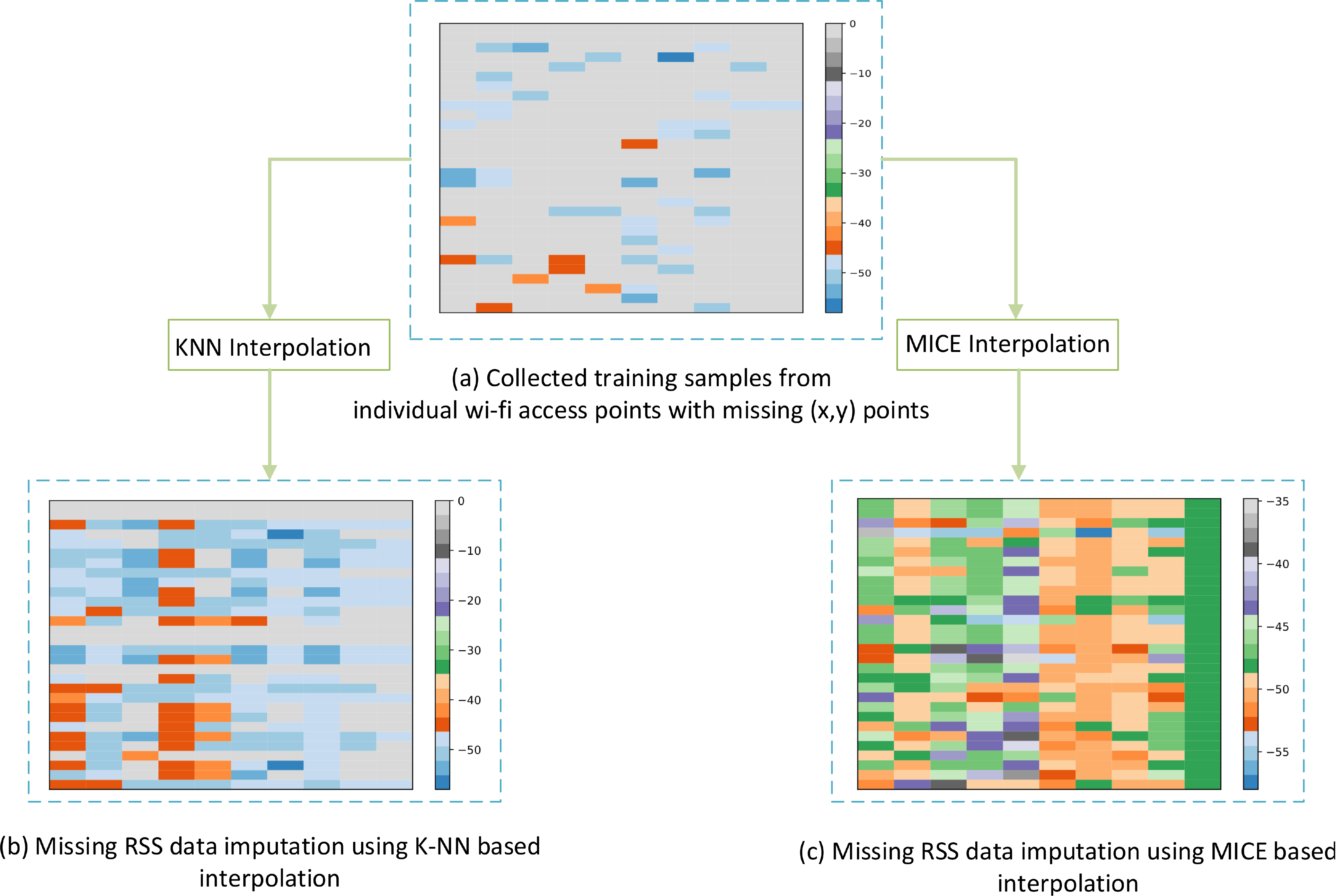}
    \caption{Interpolation of RFMap data using $k$-NN and MICE interpolation techniques. Due to limited amount of measurement samples, performance of $k$-NN is sub-optimal compared to MICE. }
    \label{fig:interpol}
    \vspace{-0.35cm}
\end{figure}

In this paper, we have employed $k$-NN and MICE for spatial interpolation of RSS data. $k$-NN is often used for Wi-Fi fingerprinting~\cite{zhao2016applying} and provides sufficed accuracy in a static channel environment. In our dataset, since the number of measurements samples were small, it was not able to perform imputation for all the missing values. For our particular data, we had a large amount of missing data and the distribution is random. From our experiment, we found that MICE yielded the best performance. The positions we chose for data collection were drawn from a Poisson Point Process (PPP) spatial distribution, which is a practical approach for creating datasets based on a set of imputation models, the one model for each variable possessing missing values. It should be noted that MICE is an increasingly popular method of performing multiple imputation~\cite{royston2011multiple}. For our dataset, MICE gave us the lowest error with respect to missing-data imputation, thus we chose this model for dataset interpolation for the GAN.

Localization is performed by a multi-layer perceptron (MLP) in our three-way DNN model, which is commonly used for feed-forward neural networks in practice. The generated data produced by the GANs model is passed to our MLP for training using the \emph{X} and \emph{Y} location values. Due to the limited amount of data, we employed the $90$/$10$ data split (\emph{i.e.}, $90$\% of the data was used as training set whereas the remaining $10$\% was used for validation and computing the mean squared error (MSE)). 

\begin{figure}[!ht]
    \centering
    \includegraphics[width=0.475\textwidth]{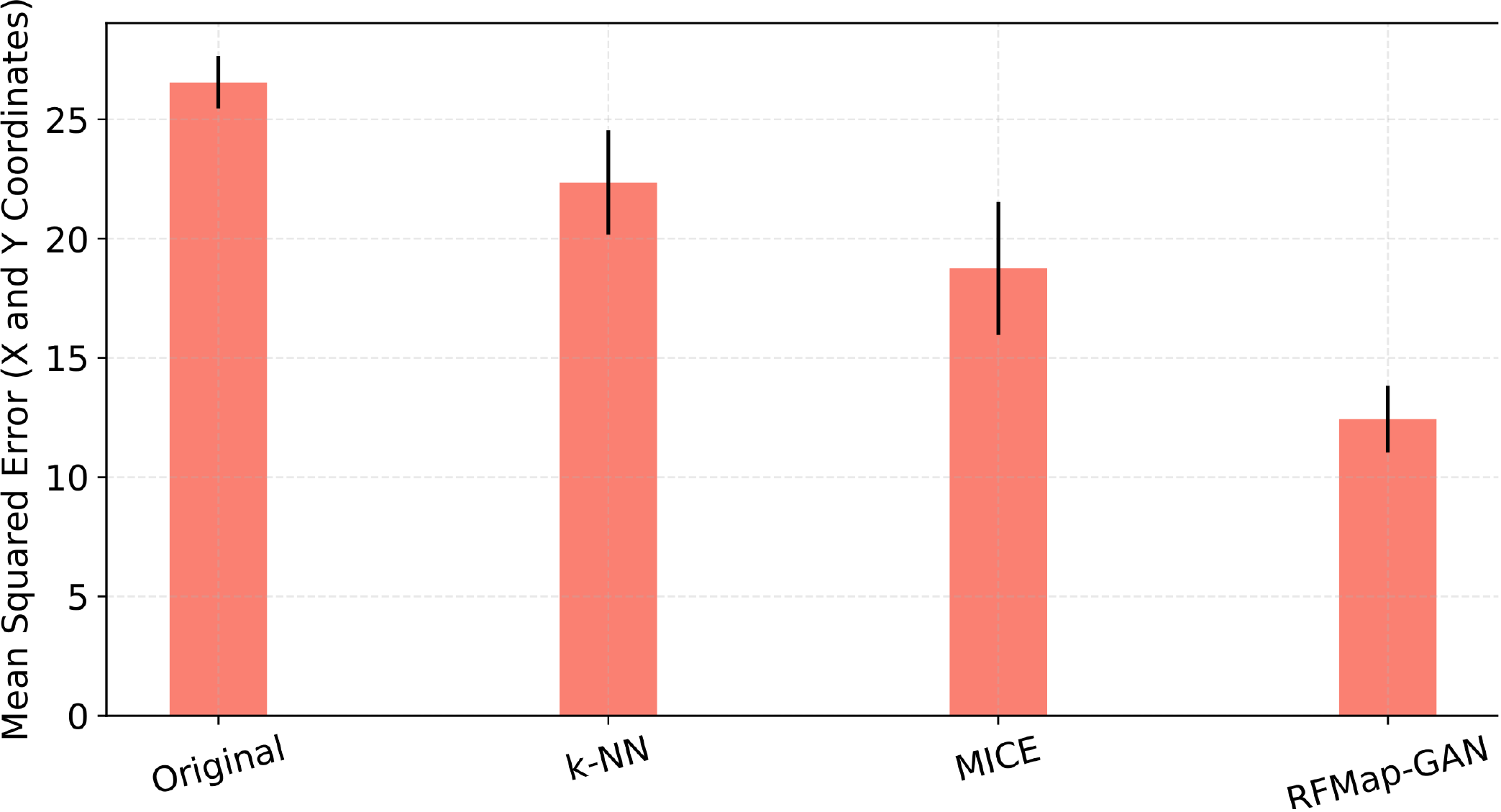}
    \caption{Localization Error for different techniques employed for missing data imputation for ten independent runs. Missing data and the interpolated data through $k$-NN, MICE and GAN are evaluated using a MLP neural network and localization errors are computed.}
    \label{fig:ganimputation}
    \vspace{-0.65cm}
\end{figure}

Finally, we computed the MSE error using our three-way DNN model and compared it against three baseline models which are as follows:
\begin{itemize}
    \item Original: This dataset possessed missing values which was directly used with MLP and the MSE error was computed. 
    \item $k$-NN: In second baseline model, we interpolated the training data using $k$-NN and then applied the MLP. 
    \item MICE: For our final baseline model, we use MICE interpolation and compute the MSE.
\end{itemize}
Figure~\ref{fig:ganimputation} shows the mean-squared error (MSE) for $X$ and $Y$ regression and its performance is compared against the three different techniques explained above. The error bar on each model represents the standard deviation after $10$ runs. The original dataset performs the worst as it had lot of missing data samples and no imputation was performed. Using $k$-NN interpolation, we see some improvement as the MSE improves but it still underperforms compared to MICE interpolation. Our proposed method performs better than other models and has the lowest MSE. The result demonstrates the robustness of our proposed approach and how it can be employed for accurate source localization in multipath environment.

\section{Conclusion and Future Work}
\label{sec:concl}

In this work, we have implemented a three-way deep neural network to perform localization. A generative adversarial network was used to construct a smooth RFMap dataset after being trained on with MICE-interpolated ground truth. The localization mean-squared error was compared for missing data, $k$-NN and MICE interpolated data and finally on GAN data and the performance for GAN was observed to be the best out of these techniques.

\section*{Acknowledgment}
The authors would like to acknowledge the guidance and input from Prof. Kaveh Pahlavan for this paper.
\bibliographystyle{IEEEtran}
\bibliography{references}
\end{document}